\pdfoutput=1

\documentclass[11pt]{article}

\usepackage[final]{acl}

\usepackage{times}
\usepackage{latexsym}

\usepackage[T1]{fontenc}

\usepackage[utf8]{inputenc}

\usepackage{microtype}

\usepackage{inconsolata}

\usepackage{graphicx}

\usepackage{booktabs}
\usepackage{makecell}
\usepackage{tabularx}
\usepackage{enumitem}
\usepackage[dvipsnames]{xcolor}
\usepackage{fancyvrb}
\usepackage{float}

\newcommand{\rd}[1]{\textcolor[rgb]{0.75,0,0}{#1}}

\title{Augmenting Dialog with Think-Aloud Utterances for Modeling Individual Personality Traits by LLM}

\author{
 \textbf{Seiya Ishikura\textsuperscript{1}},
 \textbf{Hiroaki Yamada\textsuperscript{1}},
 \\
 \textbf{Tatsuya Hiraoka\textsuperscript{2,3}},
 \textbf{Hiroaki Yamada\textsuperscript{4}},
 \textbf{Takenobu Tokunaga\textsuperscript{1}}
\\
\\
 \textsuperscript{1}Institute of Science Tokyo,
 \textsuperscript{2}Mohamed bin Zayed University of Artificial Intelligence,
\\
 \textsuperscript{3}Nara Institute of Science and Technology,
 \textsuperscript{4}Fujitsu Limited
\\
 \small
 \texttt{ishikura.s.1771@m.isct.ac.jp},
 \texttt{yamada@comp.isct.ac.jp},
 \\
 \small
 \texttt{tatsuya.hiraoka@mbzuai.ac.ae},
 \texttt{yamadah@fujitsu.com},
 \texttt{take@c.titech.ac.jp}
}

\begin{document}
\maketitle
\begin{abstract}
This study proposes augmenting dialog data with think-aloud utterances (TAUs) for modeling individual personalities in text chat by LLM.
TAU is a verbalization of a speaker's thought before articulating the utterance.
We expect ``persona LLMs'' trained with TAU-augmented data can mimic the speaker's personality trait better.
We tested whether the trained persona LLMs obtain the human personality with respect to Big Five, a framework characterizing human personality traits from five aspects.
The results showed that LLMs trained with TAU-augmented data more closely align to the speakers' Agreeableness and Neuroticism of Big Five than those trained with original dialog data. 
We also found that the quality of TAU-augmentation impacts persona LLM's performance.
\end{abstract}

\section{Introduction}
Tuning LLMs to replicate specific personas is an active research field.
Persona is a particular personality or a set of traits that characterize an individual. 
A model aligned with a persona can facilitate consistent interactions with users~\cite{zhang-etal-2018-personalizing} and enable applications in various industries, including entertainment~\cite{yan2023larplanguageagentroleplay}.

Previous research on persona replication targeted celebrities~\cite{shao-etal-2023-character, wang-etal-2024-rolellm, li2023chatharuhirevivinganimecharacter}, as their biographies and profiles are easily available online.
For instance, \citet{shao-etal-2023-character} replicated historical figures and fictional characters by training LLMs to reproduce their thought and utterance styles.
They synthesize dialogs with verbalized target person's thought using external knowledge, such as Wikipedia. Then, they fine-tune the model using the synthesized dialogs.
In contrast, replicating non-celebrities' personas remains challenging because their detailed profiles are rarely available.
Some studies employ techniques such as manipulating neurons and activations to bring LLM responses closer to the target persona~\cite{deng2025neuron, zhu2025personality}.
\citet{li-etal-2025-big5} embedded human personality traits into LLMs by training them on a large-scale dialog dataset called BIG5-CHAT. They also employed psychological tests to evaluate the models' personality traits.

Inspired by \citet{shao-etal-2023-character}, we augment real dialog data to include non-celebrities without their personal information; dialog data is easier to obtain for non-celebrities than their personal information. 
We propose augmenting the dialogs with their think-aloud utterances (TAUs).
A TAU is a verbalization of the speaker's internal psychological states, such as emotions and thoughts, before articulating the utterance.
It is widely used in cognitive science to study human cognitive processes~\cite{10.7551/mitpress/5657.001.0001}.
Our method automatically inserts a TAU before each utterance using LLM since it is difficult to collect TAUs from human interlocutors directly. 
We hypothesize that TAUs contain richer information about human personality than surface utterances; thus, training with the TAU-augmented dialog data can make the persona LLM closer to the target individual's personality.

\begin{figure}[t]
    \centering
    \includegraphics[width=\linewidth]{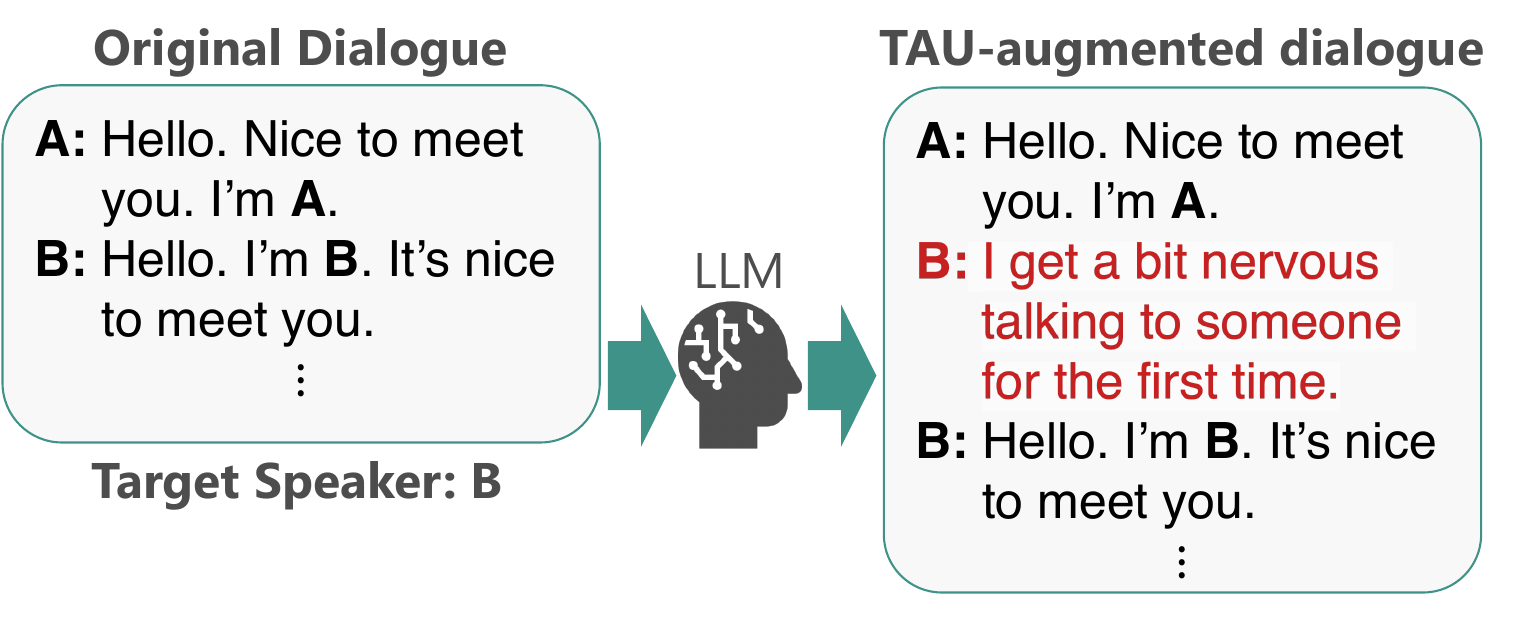}
    \caption{Augmenting dialog data with think-aloud uttrances (TAU; indicated in red) generated by LLM.} 
    \label{fig:add_think_aloud}
\end{figure}

For evaluation, we apply a psychological test that assesses human personality in terms of the Big Five aspects  (Openness, Conscientiousness, Extraversion, Agreeableness, Neuroticism)~\cite{goldberg, McCrae} to the trained LLMs.
Previous research used similar psychological methodologies to investigate the personality traits embedded within LLMs for a deeper understanding of model behavior~\cite{jiang-etal-2024-personallm, song2023largelanguagemodelsdeveloped, frisch-giulianelli-2024-llm}. 

The experiment compares the LLM fine-tuned with the original dialog data and that with the TAU-augmented data.
Our findings are (a) the latter shows better personality conformity with individuals in  Agreeableness and Neuroticism of the Big Five, and (b) the quality of TAU augmentation impacts LLMs' performance.

\section{Proposed Method}
The proposed method comprises the following two steps: (i) augmenting dialogs with TAU and (ii) fine-tuning with the TAU-augmented dialog.

\subsection{TAU Augmentation}
We use LLMs to add a TAU before each target speaker's utterances in dialog data, as shown in Figure~\ref{fig:add_think_aloud}.
Ideally, dialog data annotated with TAUs would be used for fine-tuning LLMs, but such data is not available, and creating such data from scratch is expensive and time-consuming.
Given a dialog with a specified target speaker (Speaker B in Figure~\ref{fig:add_think_aloud}), the LLM returns the augmented dialog with the target speaker's TAU.
Table~\ref{tab:augmentation-prompt} in Appendix~\ref{apx:prompts} shows the prompt for the TAU augmentation.
Table~\ref{tab:example_think_aloud_with_BigFive} presents a dialog example augmented with Speaker B's TAUs, who has a high Neuroticism score of Big Five.

A person with high Neuroticism often feels nervousness and anxiety~\cite{lahey2009public}.
In the example, the TAUs are aligned with the speaker's personality, where (8) and (11) describe anxiety about the future and health.

\begin{table}[t]
    \centering
    \small
    
    \begin{tabular}{r@{ }c@{ }p{0.7\linewidth}}
    \toprule
    \textbf{No.}& \textbf{Speaker}& \quad \textbf{Utterances (TAUs in red)}\\
    \midrule
    (1) & A& Nice to meet you.\\
    (2) & \rd{B}& \rd{\textit{First, I need to respond with a polite greeting.}}\\
    (3) & B& Nice to meet you.\\
    (4) & A& Is there anything you'd like to say to yourself five years from now?\\
    (5) & \rd{B}& \rd{\textit{Five years from now, huh… I’ve thought about it before, but picturing my current self is really difficult.}}\\
    (6) & B& I'd want to ask, ``Are you working properly?''\\
    (7) & A& Are you not working at the moment?\\
    (8) &\rd{B}& \rd{\textit{I’m still enjoying student life, but I feel anxious about the future.}}\\
    (9) & B& I'm in college now, so not yet.\\
    (10) & A& I see. I'm sure you'll be doing just fine at work by then.\\
    (11) & \rd{B}& \rd{\textit{I hope so. But I’m also worried about my health.}}\\
    (12) & B& I just hope I won't be depressed.\\
    & \vdots &\\
    \bottomrule
    \end{tabular}
    \caption{Dialog example with Speaker B's TAUs. (Translation from the original dialog in Japanese)\label{tab:example_think_aloud_with_BigFive}}

\end{table}

\subsection{Fine-tuning with TAU-augmented Dialog}
By fine-tuning the persona LLM using dialog data augmented with a target speaker's TAUs, we aim to create the LLM  mimicking the target speaker. 
An augmented dialog is converted into a multi-turn prompt~(Appendix~\ref{apx:multiturn-prompt}, Table~\ref{tab:think_aloud_messages}).
Each turn of the target speaker consists of a TAU followed by its original utterance in the format of ``\texttt{<thinking>} \{TAU\} \texttt{</thinking>} \{original utterance\}'', where we enclose TAU with the \texttt{<thinking>} tags\footnote{Throughout the paper, strings enclosed by curly blackets denote a place holder to be filled in the actual prompt.}.
We assign the \texttt{assistant} role to the target speaker and the \texttt{user} role to their counterpart, so that the LLM can distinguish the interlocutors of the dialog.

We train the model to generate the target speaker's last TAU and corresponding utterance.
The model receives all preceding utterances up to immediately before the target speaker's last turn and is trained to predict their last turn, i.e., TAU and the utterance.
The model predicts all target speaker's turns in a dialog during the training.
We implemented our fine-tuning experiment with the \texttt{TRL}~\citep{vonwerra2022trl} library.

\section{Experiment}
\subsection{Dataset}
We utilize RealPersonaChat (RPC)~\cite{yamashita-etal-2023-realpersonachat}, which is a Japanese corpus of casual text chat dialogs.
The corpus includes the interlocutors' demographic information and personality traits, which are represented using the Big Five framework.
Each trait of the Big Five is scored on a scale of 1 to 7.
Twenty individuals are randomly selected as our targets, provided they have participated in at least 190 dialogs.
Their dialogs, 3,929 in total, are used for our experiments. 
We splited the data into training, validation and test sets at an approximate ratio of $8:1:1$ for each speaker.

\subsection{TAU Augmentation}
We use Qwen2.5-72B-Instruct~\cite{qwen2025qwen25technicalreport} and gpt-4o-2024-08-06 (gpt-4o)~\cite{openai2024gpt4ocard} for TAU augmentation.

An augmented dialog must consist of multiple utterances separated by new lines. An utterance is either an original utterance or a TAU.
Also, a TAU must be inserted immediately before its target speaker’s corresponding original utterance. 
Table~\ref{tab:TAU-augmentation-format} in Appendix~\ref{apx:format} shows an example output of TAU augmentation. 
If an LLM produces an output that violates the specified format, we repeat the TAU generation until the output complies with the format.
If it still fails after 15 iterations, we discard the dialog from both the original RPC training data and the augmented training data.
In our experiment, we observed no such case from gpt-4o, and eight out of 2,929 cases from Qwen2.5-72B-Instruct.

In the augmentation, the LLMs output the TAUs along with their original utterances from the inputs. As LLMs can introduce slight modifications in the original utterances in their outputs, we simply replace the corresponding utterances with the original utterances in post-processing if necessary, thereby maintaining consistency between the original utterances before and after augmentation.

\subsection{Models to Compare}
We compare the following models to evaluate the effectiveness of TAU-augmentation for predicting the target speaker's personality traits. Bare LLMs are used as \textbf{Base} model, which represent the LLMs' personality traits before any training.
The \textbf{w/ RPC} model is trained with the original RPC dialogs, while the \textbf{w/ +TAU} model is trained with their TAU-augmented version.

We employed four base LLMs: (1) gpt-4o-mini-2024-07-18 (gpt-4o-mini), (2) Llama-3-Swallow-8B-Instruct-v0.1~\cite{Fujii:COLM2024, Okazaki:COLM2024}, (3) Qwen2.5-7B-Instruct and (4) gemma-2-9b-it~\cite{gemmateam2024gemma2improvingopen}.
We fine-tuned gpt-4o-mini via the OpenAI API. 

Due to the license constraints, the data augmented by gpt-4o is only for training gpt-4o-mini (1).
We use QLoRA~\cite{dettmers2023qloraefficientfinetuningquantized} and the augmented data by Qwen2.5-72B-Instruct for fine-tuning all the other models (2, 3, 4). The \texttt{lora\_rank} and \texttt{lora\_alpha} hyperparameters are set to 64.
We choose the learning rate for each model according to the similarity scores (Appendix~\ref{sec:utterance_eval}) calculated between the generated utterances and their references in the validation data.

\section{Evaluation}
We assess the model's personality traits using a psychological test that measures the Big Five traits~\citep{wada}.
As RPC has also assigned the Big Five scores to the dialog participants, 
we can evaluate the model's ability to reproduce the target speaker's traits by comparing the Big Five scores from the models and those from RPC.

The test consists of 60 items, each 12 items related to one of the Big Five traits.
The responses to the items are provided on a 7-point Likert scale; they are averaged over 12 items of the corresponding Big Five trait to give its score.
We input each question item to the models one by one and let the models generate an answer.
We expect the models to start their responses with one of the scale points or labels associated with the points.
If the models do not start with either of them, we will select the highest probability token representing the scale points at the beginning of utterances.

As the evaluation metric, we use mean squared error (MSE) between the model's estimated scores and the speaker's scores of each Big Five trait.
A smaller MSE indicates that the model more accurately reflects the speaker's personality traits.

Variations in prompting may cause fluctuations in model responses~\citep{gupta-etal-2024-self}.
To reduce response biases caused by differences in prompts, we combine three factors to create eight variations of each prompt; three factors are (1) the order of the question sentence and response scale, (2) the scale point labels, numeric vs Roman alphabets and (3) the scale order, from ``strongly disagree'' to ``strongly agree'' or vice versa.
We average the MSEs obtained from the eight prompt variations for each Big Five factor.

\section{Results and Discussions}
\begin{table}[t]
    \centering
    \small
        \begin{tabularx}{\linewidth}{lXXXXXX}
        \toprule
        \textbf{Model} & \textbf{O} & \textbf{C} & \textbf{E} & \textbf{A} & \textbf{N} & \textbf{Mean}\\
        \midrule
        \multicolumn{7}{l}{[\textit{gpt-4o-mini}]} \\[1ex]
        Base & 1.557 & 1.262 & 1.771 & 0.658 & 1.763 & 1.402 \\
        w/ RPC & \textbf{0.975} & \textbf{1.048} & \textbf{1.505} & 0.670 & 1.662 & \textbf{1.172}  \\
        w/ +TAU & 1.088 & 1.107 & 1.583 & \textbf{0.630} & \textbf{1.571} & 1.196  \\
        \midrule
        \multicolumn{7}{l}{[\textit{Llama-3-Swallow-8B-Instruct-v0.1}]}\\[1ex]
        Base  & \textbf{0.729} & 1.070 & \textbf{1.567} & \textbf{0.861} & 1.971 & \textbf{1.240} \\
        w/ RPC & 0.919 & \textbf{1.061} & 1.635 & 1.309 & 1.937 & 1.372 \\
        w/ +TAU & 0.882 & 1.178 & 1.705 & 1.267 & \textbf{1.855} & 1.377  \\
        \midrule
        \multicolumn{7}{l}{[\textit{Qwen2.5-7B-Instruct}]} \\[1ex]
        Base  & \textbf{0.996} & \textbf{1.227} & 1.471 & \textbf{0.525} & \textbf{1.823} & \textbf{1.208} \\
        w/ RPC & 1.045 & 1.351 & \textbf{1.453} & 0.740 & 2.377 & 1.393 \\
        w/ +TAU & 1.124 & 1.336 & 1.562 & 0.643 & 1.936 & 1.320 \\
        \midrule
        \multicolumn{7}{l}{[\textit{gemma-2-9b-it}]}  \\[1ex]
        Base  & 1.091 & \textbf{0.925} & 1.988 & 1.023 & 1.977 & 1.401 \\
        w/ RPC & 0.964 & 0.935 & 1.810 & 0.891 & 1.800 & 1.280  \\
        w/ +TAU & \textbf{0.874} & 0.989 & \textbf{1.702} & \textbf{0.749} & \textbf{1.745} & \textbf{1.212} \\
        \bottomrule
        \end{tabularx}
        \caption{MSEs between the models' estimated and true scores of the Big Five traits. Bold face indicates the minimum MSE in the base LLM.\label{tab:bigfive_dim_think_aloud_plane}}
\end{table}
Table~\ref{tab:bigfive_dim_think_aloud_plane} presents the MSEs between the models' estimated scores of the Big Five traits and those from RPC.
The model with the minimum mean value (in bold face) varies across the base LLMs.
The TAU-augmentation is effective only with gemma-2-9b-it.
However, the w/ +TAU model consistently shows smaller MSEs than the w/ RPC model for Agreeableness (A) and Neuroticism (N), although they are not always the minimum across all base LLMs.
This indicates that the TAU-augmentation helps the model to learn these two traits from the dialog data. 

On the other hand, we observe more inconsistent results on Openness (O), Conscientiousness (C), and Extraversion (E). We assume that the source of this inconsistency is the characteristics of the dataset we used. The RPC consists of dialogs collected from recruited participants who were asked to engage in text-based conversations in pairs.
This artificial setting may have constrained the natural expression of personality traits observable in the collected dialogs, resulting in limited improvement by our proposed method.
Particularly, Extraversion (E) should be affected by its nature, as the participants were required to engage in active conversations, regardless of their level of extraversion.

Bare Llama-3-Swallow-8b-Instruct-v0.1 and Qwen2.5-7b-Instruct show the smaller MSEs without fine-tuning using the dialog data, regardless of the TAU augmentation.
The effectiveness of TAU augmentation is limited regarding both base LLMs and Big Five traits.

\begin{table}[t]
    \centering
    \small
    \tabcolsep 2pt
    \begin{tabularx}{\linewidth}{lXXXX}
    \toprule
    \textbf{Model} & \textbf{\textit{gpt-4o-mini}} & \textbf{\textit{Llama-3-Swallow}} & \textbf{\textit{Qwen2.5}} & \textbf{\textit{gemma-2}}\\
    \midrule
    w/ RPC& -0.614 & -0.013 & -0.389 & -0.032 \\
    & (0.004) & (0.956) & (0.090) & (0.892) \\
    w/ +TAU& -0.693 & -0.144 & -0.130 & -0.182 \\
    & (0.001) & (0.544) & (0.586) & (0.441) \\
    \bottomrule
    \end{tabularx}
    \caption{Correlation coefficients between base model MSE and the gains of w/ RPC and w/ +TAU over base model (p-values in the parentheses.)\label{tab:corr_coef_between_Base_and_gain}}
\end{table}

\begin{table}[t]
    \centering
    \small
    \tabcolsep 4pt
        \begin{tabular}{lrrrrrr}
        \toprule
        \textbf{LLM} & \multicolumn{1}{c}{\textbf{O}} & \multicolumn{1}{c}{\textbf{C}} & \multicolumn{1}{c}{\textbf{E}} & \multicolumn{1}{c}{\textbf{A}} & \multicolumn{1}{c}{\textbf{N}} & \multicolumn{1}{c}{\textbf{Mean}}\\
        \midrule
        Qwen& 0.116 & 0.301 & -0.020 & -0.011 & 0.310 & 0.139   \\
        gpt & 0.113 & 0.059 & 0.077 & -0.040 & -0.091 & 0.024 \\
        gpt+BF& 0.278 & 0.101 & -0.261 & -0.087 & -0.238 & -0.041  \\
        \bottomrule
        \end{tabular}
        \caption{MSE gains of w/+TAU from w/RPC of gpt-4o-mini per TAU-augmentation LLM (Qwen, gpt, gpt+BF).\label{tab:diff_TAU_and_RPC}}
\end{table}

We further investigate the individual-wise effectiveness of fine-tuning.
Table~\ref{tab:corr_coef_between_Base_and_gain} shows the Pearson correlation coefficients~\citep{PearsonVIINO} between the MSEs of the base model and the gains of the fine-tuned models (w/ RPC and w/ +TAU) over the base model.
The gains are calculated by subtracting the base models' MSEs from the MSEs of the fine-tuned models.
The negative gains indicate improvement from the base models.
Both w/ RPC and w/ +TAU in gpt-4o-mini exhibit a negative correlation, which indicates that the fine-tuning is more effective on the individuals whose personalities are distant from the base model's Big Five.

Next, we investigate how TAUs generated by different LLMs affect model training.
We compare two variants of gpt-4o-mini, each fine-tuned on TAU-augmented dialogs, where the TAUs are generated by either Qwen2.5-72B-Instruct or gpt-4o.
Table~\ref{tab:diff_TAU_and_RPC} shows the gains of MSE by TAU-augmentation, i.e. MSE of w/ +TAU minus MSE of w/ RPC, for the LLMs used for augmentation.
In Table~\ref{tab:diff_TAU_and_RPC}, a negative value indicates that TAU helps the models learn personality traits, and a smaller value suggests more improvement.
The table shows TAU-augmentation by gpt-4o (gpt) provides better TAUs compared with Qwen2.5-72B-Instruct (Qwen), except for Extraversion (E).

We create another TAU-augmented dialog by providing gpt-4o with the target speaker's Big Five scores and descriptions of Big Five in the system prompt (``gpt+BF'' in Table~\ref{tab:diff_TAU_and_RPC}). Table~\ref{tab:augmentation-prompt-with-BF} in Appendix~\ref{apx:prompts} shows the prompt used in TAU augmentation with Big Five.
We expect that the input Big Five scores are reflected in the generated TAUs and improve TAU-augmentation quality.
Providing the Big Five scores can be indirect cheating; we restrict their use to testing the impact of TAU quality on personality learning.
We observe further improvement by TAU augmentation in the Extraversion (E), Agreeableness (A) and Neuroticism (N), but not in the Openness (O) and Conscientiousness (C).
We conclude that the quality of TAU-augmentation impacts the performance of persona LLMs.

\section{Conclusion}
We proposed fine‑tuning LLMs with TAU-augmented dialogs to learn a speaker's personality.
The experiments showed that TAU-augmentation is effective for LLMs to learn Agreeableness and Neuroticism of the Big Five traits.
Also, we showed that the effectiveness of TAU augmentation varies among speakers' personality traits and the TAU quality.

In future work, we plan to construct a new dataset of dialogs with real TAUs by asking human interlocutors to verbalize their thoughts and feelings during their conversations.
Using this dataset, we assess the quality of TAUs by comparing the model-generated TAUs against the real ones. 
Furthermore, we will expand our experiments by incorporating a wider variety of dialogues across different scenarios with different types of participant pairs (e.g., friends, family, strangers) to confirm the generalizability of our method.

\section*{Limitations}
We use RealPersonaChat which is a corpus of text-chat dialogs collected from crowdsourcing workers who have never met each other.
Our proposed methods utilize utterances to infer personality traits and generate TAUs. Thus, our methods can behave differently when we use dialogs collected from speakers with different relationships.
This limited scope of interpersonal relationships may affect the generalizability of our findings.

Additionally, due to the license constraints, TAUs augmented by gpt-4o are applied only to fine-tune gpt-4o-mini. Consequently, we could not train non-OpenAI models with the TAUs from gpt-4o. 
Thus, we could not explore whether the gpt-4o-generated TAUs could further improve the performance of the models, such as Llama-3-Swallow-8B-Instruct-v0.1, Qwen2.5-7B-Instruct and gemma-2-9b-it.

\section*{Acknowledgments}
This work was supported by JST PRESTO Grant Number JPMJPR236B and JPMJPR236C.

\bibliography{anthology,custom}

\clearpage
\appendix
\section*{Appendix}

\section{Models}
\label{apx:models}
We list all LLMs employed in our experiments.
\subsection{TAU augmentation}
\label{apx:TAU-augmentation}
\begin{itemize}[leftmargin=*, label={}, itemsep=0ex,]
    \item Qwen2.5-72B-Instruct (\url{https://huggingface.co/Qwen/Qwen2.5-72B-Instruct}) via deepinfra.com API
    \item  gpt-4o-2024-08-06 (\url{https://platform.openai.com/docs/models/gpt-4o}) via openai.com API
\end{itemize}

\subsection{Fine-tuning}
\label{apx:Fine-tuning}
\begin{itemize}[leftmargin=*, label={}, itemsep=0ex,]
    \item Llama-3-Swallow-8B-Instruct-v0.1 (\url{https://huggingface.co/tokyotech-llm/Llama-3-Swallow-8B-Instruct-v0.1}) via our GPU server
    \item Qwen2.5-7B-Instruct (\url{https://huggingface.co/Qwen/Qwen2.5-7B-Instruct}) via our GPU server
    \item gemma-2-9b-it (\url{https://huggingface.co/google/gemma-2-9b-it}) via our GPU server
    \item gpt-4o-mini-2024-07-18 (\url{https://platform.openai.com/docs/models/gpt-4o-mini}) via openai.com API
\end{itemize}

\subsection{Our GPU server specifications}
\label{apx:gpu-spec}
\begin{itemize}[leftmargin=*, label={}, itemsep=0ex,]
    \item CPU: AMD EPYC 9654 2.4GHz × 2 Socket
    \item RAM: 768GiB
    \item GPU: NVIDIA H100 SXM5 94GB HBM2e × 4*
    *Only one GPU was used in our experiments.
\end{itemize}

\section{Prompt for TAU augmentation\label{apx:multiturn-prompt}}
\label{apx:prompts}
Table~\ref{tab:augmentation-prompt} shows the prompt used in TAU augmentation.

\begin{table}[t]
    \centering
    \small
    \begin{tabular}{p{0.94\linewidth}}
    \toprule
        \textbf{System prompt}\\
        \midrule
        \#\# Basic Information\\
        You are \{target\_interlocutor\_id\}. \\
        \specialrule{1pt}{0pt}{2pt}
        \textbf{User prompt}\\
        \midrule
        \#\# Task Description\\
	1.	Add a think-aloud utterance for \{target\_interlocutor\_id\}. The think-aloud utterance should express what \{target\_interlocutor\_id\} is thinking and feeling.\\
	2.	Insert the think-aloud utterance on a line that starts with ``\{target\_interlocutor\_id\} (thinking):''.\\
        \\
        \#\# Format Description\\
        Below are examples of the original dialog history and the required output format.\\
        \\
        Dialog History Example 1:\\
        \{utterances\_example\_1\}\\
        Output Example 1:\\
        \{think\_aloud\_utterances\_example\_1\}\\
        \\
        Dialog History Example 2:\\
        \{utterances\_example\_2\}\\
        Output Example 2:\\
        \{think\_aloud\_utterances\_example\_2\}\\
        \\
        When generating your answer, do not include the label ``Output:'' and do not insert blank lines in the output.\\
        \\
        \#\# Task\\
        Read the dialog history below and add think-aloud utterance for \{target\_interlocutor\_id\}.\\
        \\
        Dialog History:\\
        \{original\_utterances\}\\
        Output:\\
    \bottomrule
    \end{tabular}
    \caption{Prompt for TAU augmentation. The original prompt is written in Japanese.  \label{tab:augmentation-prompt}}
\end{table}

\begin{table}[t]
    \centering
    \small
    \begin{tabular}{p{0.94\linewidth}}
    \toprule
        \textbf{System prompt}\\
        \midrule
        \#\# Basic Information\\
        You are \{target\_interlocutor\_id\}. Your Big Five personality traits are as follows.\\
        Big Five Personality Traits (on a scale from 1 to 7, where 1 is the minimum and 7 is the maximum): \{target\_interlocutor\_personality\_prompt\}\\
        Openness reflects the degree of interest in and curiosity about new ideas and experiences. People high in openness tend to be creative, imaginative, and eager to engage with new things.\\
        Conscientiousness indicates the degree of self-discipline, responsibility, and motivation toward achieving goals. Those high in conscientiousness are reliable and have strong self-management skills.\\
        Extraversion represents the level of sociability, activity, and positive emotion. Highly extraverted individuals are sociable, energetic, and often enjoy interactions with others.\\
        Agreeableness shows the degree of consideration, cooperation, and empathy toward others. People high in agreeableness value relationships and are sensitive to others’ feelings and needs.\\
        Neuroticism reflects the level of anxiety, depression, and stress tolerance. Individuals high in neuroticism tend to experience stronger emotional fluctuations and lower resistance to stress.\\
    \specialrule{1pt}{0pt}{2pt}
    \textbf{User prompt}\\
        \midrule
        \#\# Task Description\\
	1.	Add a think-aloud utterance for \{target\_interlocutor\_id\}. The think-aloud utterance should express what \{target\_interlocutor\_id\} is thinking and feeling.\\
	2.	Insert the think-aloud utterance on a line that starts with ``\{target\_interlocutor\_id\} (thinking):''.\\
    3.  Ensure that the Big Five personality traits of \{target\_interlocutor\_id\} are always reflected in the think-aloud utterance.\\
        \\
        \#\# Format Description\\
        Below are examples of the original dialog history and the required output format.\\
        \\
        Dialog History Example 1:\\
        \{utterances\_example\_1\}\\
        Output Example 1:\\
        \{think\_aloud\_utterances\_example\_1\}\\
        \\
        Dialog History Example 2:\\
        \{utterances\_example\_2\}\\
        Output Example 2:\\
        \{think\_aloud\_utterances\_example\_2\}\\
        \\
        When generating your answer, do not include the label ``Output:'' and do not insert blank lines in the output.\\
        \\
        \#\# Task\\
        Read the dialog history below and add think-aloud utterance for \{target\_interlocutor\_id\}.\\
        \\
        Dialog History:\\
        \{original\_utterances\}\\
        Output:\\
    \bottomrule
    \end{tabular}
    \caption{Prompt for TAU augmentation with Big Five scores and descriptions. The original prompt is written in Japanese.  \label{tab:augmentation-prompt-with-BF}}
\end{table}

\section{Multi-turn Prompt for Fine-tuning}
Table~\ref{tab:think_aloud_messages} shows an example of multi-turn style TAU-augmented data used for fine-tuning LLMs.

\begin{table}[t]
    \centering
    \small
    \begin{tabular}{l@{\quad}p{0.75\linewidth}}
    \toprule
    \textbf{Role} & \textbf{Content} \\
    \midrule
    user & Nice to meet you  \\
    assistant & \rd{\texttt{<thinking>}\textit{First, I need to respond with a polite greeting.}\texttt{</thinking>}} Nice to meet you.\\
    user & Is there anything you'd like to say to yourself five years from now?\\
    assistant & \rd{\texttt{<thinking>} \textit{Five years from now, huh… I’ve thought about it before, but picturing my current self is really difficult. }\texttt{</thinking>}} I'd want to ask, ``Are you working properly?''\\
    \multicolumn{2}{c}{\vdots}\\
    \bottomrule
    \end{tabular}
    \caption{Example of multi-turn prompt with TAUs. (Translation from the original dialog in Japanese)\label{tab:think_aloud_messages}}
\end{table}

\section{TAU-augmented Output Format}\label{apx:format}
Table~\ref{tab:TAU-augmentation-format} shows an example of output format from TAU augmentation. 
\begin{table}[t]
    \centering
    \small
    
    \begin{tabular}{p{0.94\linewidth}}
    \toprule
        \textbf{Model input (filled to \{original\_utterances\})}\\
        \midrule
        A (speaking): Nice to meet you.\\
        B (thinking): \\
        B (speaking): Nice to meet you.\\
        A (speaking): Is there anything you'd like to say to yourself five years from now?\\
        B (thinking): \\
        B (speaking): I'd want to ask, ``Are you working properly?''\\
        \multicolumn{1}{c}{\vdots}\\
        \specialrule{1pt}{0pt}{2pt}
        \textbf{Model output (in plain text)}\\
        \midrule
        A (speaking): Nice to meet you.\\
        B (thinking): First, I need to respond with a polite greeting.\\
        B (speaking): Nice to meet you.\\
        A (speaking): Is there anything you'd like to say to yourself five years from now?\\
        B (thinking): Five years from now, huh… I’ve thought about it before, but picturing my current self is really difficult.\\
        B (speaking): I'd want to ask, ``Are you working properly?''\\
        \multicolumn{1}{c}{\vdots}\\
    \bottomrule
    \end{tabular}
    \caption{Output format example for TAU augmentation  \label{tab:TAU-augmentation-format}}
\end{table}

\section{Definition of utterance similarity}
\label{sec:utterance_eval}
We introduce a similarity score to decide the learning rates.
For every utterance in a dialog which include target speaker from validation data, we feed the previous context into the model and have it generate the next utterance. Then we calculate BERTScore~\cite{zhang2020bertscoreevaluatingtextgeneration}, ROUGE-1~\cite{lin-2004-rouge}, ROUGE-2, and ROUGE-L between the model's utterance and the reference utterance. 
We calculate the average of each metric over all of the target speaker's utterances. We repeat this procedure for every dialog in the validation data and calculate the average of each metric across dialogs.
Finally, we define the sum of the resulting BERTScore, ROUGE-1, ROUGE-2, and ROUGE-L averages as the overall similarity score used to select the learning rate.

\end{document}